\documentclass[letterpaper, 10 pt, conference]{ieeeconf}  

\IEEEoverridecommandlockouts                              

\usepackage{graphicx} 
\usepackage{amsmath} 
\usepackage{algorithm}
\usepackage{booktabs}
\usepackage{multirow}
\usepackage{multicol}
\usepackage{algpseudocode}
\usepackage{balance}
\usepackage{placeins}
\usepackage{url}

\title{\LARGE \bf SAGE: Scene Graph-Aware Guidance and Execution \\ for Long-Horizon Manipulation Tasks}

\author{Jialiang Li$^{1}$, Wenzheng Wu$^{2}$, Gaojing Zhang$^{3}$, Yifan Han$^{4}$ and Wenzhao Lian$^{1,\dagger}$ \\
\url{}
\thanks{$^{1}$School of Artificial Intelligence, Shanghai Jiao Tong University. $^{2}$ShanghaiTech University. $^{3}$University of Sussex. $^{4}$Institute of Automation, Chinese Academy of Sciences. ${\dagger}$Corresponding Author. }
}

\begin{document}

\setcounter{figure}{1}
\makeatletter
\let\@oldmaketitle\@maketitle
\renewcommand{\@maketitle}{\@oldmaketitle
  \begin{center}
    \includegraphics[width=\textwidth]{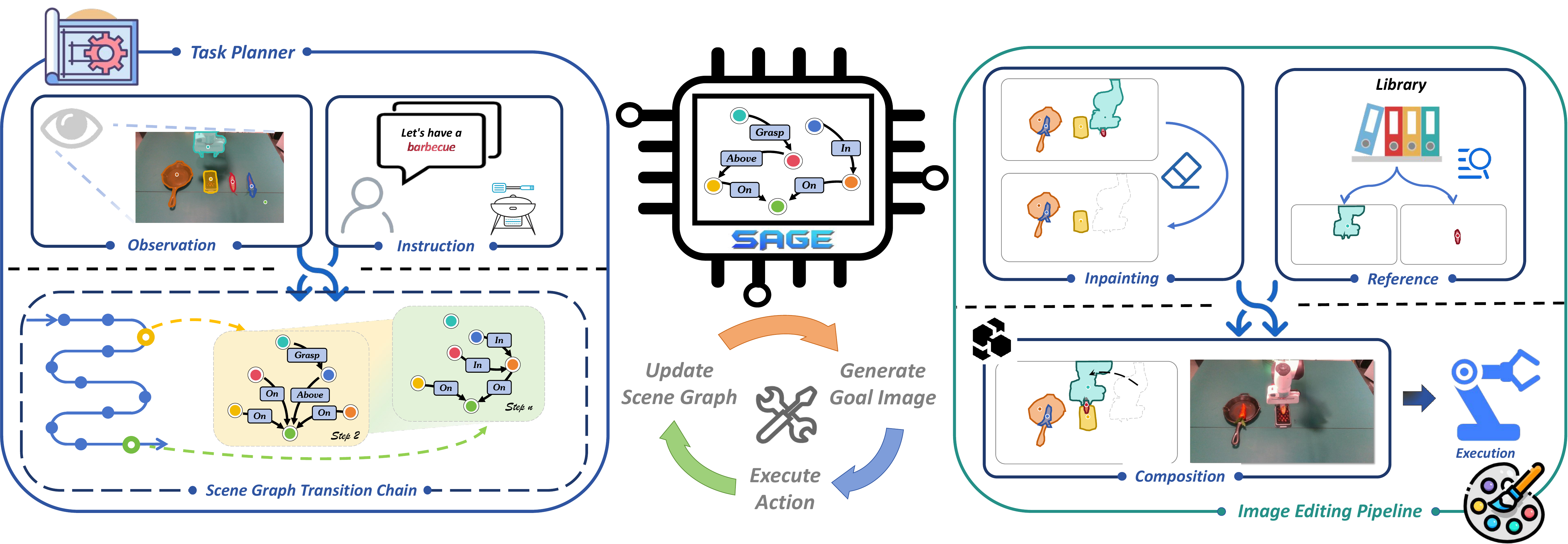}
    \label{fig:cover}
  \end{center}
  \vspace{-15pt}
  \footnotesize{Fig.~\thefigure.~\label{fig:cover} Our framework, SAGE, leverages semantic scene graphs as a structural representation for long-horizon manipulation tasks. SAGE consists of two key components: (1) a task planner that reason about a feasible and physically-grounded scene graph transition chain; and (2) an image editing pipeline that controllably converts scene graph into a sub-goal image via image inpainting and composition. The entire long-horizon manipulation task is accomplished by iteratively generating sub-goal images and executing actions.}
  \label{fig:cover}
 \vspace{-15pt}
  \medskip}
\makeatother

\maketitle
\thispagestyle{empty}
\pagestyle{empty}

\begin{abstract}
Successfully solving long-horizon manipulation tasks remains a fundamental challenge. These tasks involve extended action sequences and complex object interactions, presenting a critical gap between high-level symbolic planning and low-level continuous control. To bridge this gap, two essential capabilities are required: robust long-horizon task planning and effective goal-conditioned manipulation. Existing task planning methods, including traditional and LLM-based approaches, often exhibit limited generalization or sparse semantic reasoning. Meanwhile, image-conditioned control methods struggle to adapt to unseen tasks. To tackle these problems, we propose SAGE, a novel framework for Scene Graph-Aware Guidance and Execution in Long-Horizon Manipulation Tasks. SAGE utilizes semantic scene graphs as a structural representation for scene states. A structural scene graph enables bridging task-level semantic reasoning and pixel-level visuo-motor control. This also facilitates the controllable synthesis of accurate, novel sub-goal images. SAGE consists of two key components: (1) a scene graph-based task planner that uses VLMs and LLMs to parse the environment and reason about physically-grounded scene state transition sequences, and (2) a decoupled structural image editing pipeline that controllably converts each target sub-goal graph into a corresponding image through image inpainting and composition. Extensive experiments have demonstrated that SAGE achieves state-of-the-art performance on distinct long-horizon tasks.

\end{abstract}

\section{Introduction}

Robots are expected to work in complex, real-world environments and perform long-horizon manipulation tasks that involve multiple cascading sub-task phases \cite{feng2025reflective,luo2024multistage}. Unlike short-horizon operations, these tasks require extended action sequences, with complex dependencies between objects and states. Despite the importance, robustly solving long-horizon manipulation tasks remains a significant challenge. A key difficulty lies in bridging the semantic gap between high-level symbolic planning and low-level continuous control. For example, a high-level plan for making a cup of tea might be \texttt{put the teabag in the mug, then pour water.} While this plan is easy for humans to understand, it is too abstract to be translated into a series of precise, continuous actions, such as correctly grasping the teabag and transfering it into the mug without spilling. Successfully bridging this gap requires two capabilities: (1) generating robust long-horizon task plans that are grounded in the physical world, and (2) providing controllers with actionable goal representations that translate abstract sub-tasks into pixel-level visuo-motor commands, enabling reliable continuous control.


To address long-horizon task planning, traditional approaches like Sense-Plan-Act (SPA) \cite{kappler2018real,murphy2019introduction} and Task and Motion Planning (TAMP) \cite{garrett2021integrated,kaelbling2011hierarchical} rely on manually predefined symbolic rules and known dynamic models, which limits their application in novel real-world scenarios. A more recent line of work leveraging large language models (LLMs) \cite{huang2022language,huang2022inner,ahn2022can,rana2023sayplan} and vision-language models (VLMs) \cite{driess2023palm,rt1_brohan2022rt,huang2023voxposer,shi2024yell} face new challenges including hallucination, uncontrollable generation, and extraction of only low-level sparse semantic information from raw images.

To provide strong goal guidance, efforts have been made to explore different forms of goals, encompassing language instructions \cite{rt2_zitkovich2023rt,pi0_black2024pi_0,openvla_kim2024openvla} and images \cite{susie_black2023zero,grmg_li2025gr}. Language is much flexible for human to specify the task goals, but it lacks the precise pixel-level information needed for effective policy learning. Meanwhile, researchers have studied image-conditioned manipulation via text-to-image diffusion-based image synthesis models \cite{ho2020denoising,ip2p_brooks2023instructpix2pix}. However, due to text embeddings, being not aligned with images, these works suffer in generating high-quality, novel images for unseen tasks.

To address the above challenges, we propose \textbf{SAGE}, a \textbf{S}cene Graph-\textbf{A}ware \textbf{G}uidance and \textbf{E}xecution framework for long-horizon manipulation tasks. The key insight underlying SAGE is that many long-horizon manipulation tasks are uniquely defined by their specific temporal execution orderings and spatial object relationships. Based on this, as shown in Figure.{~\thefigure}, SAGE leverages LLMs for planning and generated images for goal guidance, while mitigating their limitations. At its core, SAGE uses semantic scene graphs \cite{semantic_chang2021comprehensive} as a structural representation of the scene state, effectively aligning task-level semantic reasoning with pixel-level visuo-motor control. A scene graph models the physical world by representing objects as nodes and their spatial relationships (e.g., \texttt{On}, \texttt{In}) as edges. We employ this representation for two reasons. First, at the planning level, the scene graphs enable LLMs to extract structural, high-level semantic information, enabling them to generate physically-grounded task plans. Second, for image-conditioned manipulation, by decomposing a scene into a disentangled scene graph with independent objects and their pairwise relations, modifications can be applied locally and consistently, allowing controllable sub-goal image generation for even unseen tasks.

SAGE consists of two key components: (1) a scene graph task planner that uses a VLM to parse the scene into a scene graph and a LLM to reason about a scene graph transition chain, which serves as a physically-grounded task plan; and (2) a decoupled structural image editing pipeline that controllably converts each scene graph of this plan into a corresponding image through image inpainting and composition. These generated images guide a visuo-motor policy to accomplish the entire long-horizon manipulation task by executing each sub-task iteratively. Experimental results have demonstrated that SAGE achieves state-of-the-art performance on various long-horizon tasks.

We summarize our contributions as follows:
\begin{itemize}
    \item We introduce SAGE, a framework that uses scene graphs to extract task keyframes, enabling the alignment between task-level semantic reasoning and pixel-level visuo-motor control for long-horizon manipulation.
    \item We develop a scene graph task planner that robustly decomposes long-horizon tasks into interpretable, physically-grounded scene graph transition chains, facilitating structural symbolic reasoning.
    \item We design a decoupled structural image editing pipeline that controllably synthesizes sub-goal images from predicted scene graphs, providing reliable visual guidance for visuo-motor control.
    \item We demonstrate that SAGE achieves SOTA performance across diverse long-horizon tasks with distinct and novel temporal execution orderings and spatial object relationships.
\end{itemize}

\section{Related Work}

\subsection{Long-Horizon Task Planning}
Traditional approaches to long-horizon task planning, such as SPA \cite{kappler2018real,murphy2019introduction,paul1981robot,whitney1972mathematics,vukobratovic1984scientific}, TAMP \cite{garrett2021integrated,kaelbling2011hierarchical,dantam2016incremental,migimatsu2020object,srivastava2014combined}, face significant limitations. They often rely on predefined symbolic rules or accurate dynamic models, which makes them difficult to generalize to complex, unknown environments. Some methods \cite{wang2021learning,driess2020deep,chane2021goal,co2018self,nachum2018near,nasiriany2019planning} require costly task-specific engineering or are computationally expensive. The progress of large language models (LLMs) \cite{huang2022language,huang2022inner,ahn2022can,rana2023sayplan,li2022pre,yao2023react} and vision language models (VLMs) \cite{driess2023palm,rt1_brohan2022rt,huang2023voxposer,shi2024yell,hu2023look,belkhale2024rt,nasiriany2024pivot} has introduced new data-driven planning paradigms. However, these models have their own challenges, they extract only the low-level sparse semantic knowledge from raw images and may incur hallucination.

Scene graphs present a promising representation that helps address these limitations. However, existing scene graph-based works have faced several challenges. Some \cite{zhu2021hierarchical,jiao2022sequential,sgbot_zhai2024sg} struggled with building geometric graphs or extracting concrete object poses, and more recent efforts \cite{grid_ni2024grid} are confined to single-object manipulation. In this paper, we leverage scene graphs to facilitate LLM-based planning for multi-object long-horizon manipulation.


\subsection{Goal-Conditioned Manipulation}
As language is intuitively comprehensive for humans, it is widely used to specify robot task goals \cite{rt2_zitkovich2023rt,pi0_black2024pi_0,openvla_kim2024openvla,LCBC_stepputtis2020language,baku_haldar2024baku}. However, a language instruction can be too abstract for effective policy learning. To provide a concrete visual grounding, recent research has focused on image-conditioned policy learning \cite{fang2024egocentric,bousmalis2023robocat,shah2023vint}. For example, \cite{susie_black2023zero,grmg_li2025gr} use diffusion models \cite{ho2020denoising,ip2p_brooks2023instructpix2pix} to generate goal images via future state prediction. However, these works often struggle to create high-quality, novel goals for unseen tasks. In this paper, we address these limitations by proposing a decoupled structural image editing pipeline, rather than a complete diffusion model, to generate controllable images representing novel scene states, which are used to condition the policy.

\begin{figure*}
    \centering    
    \includegraphics[width=\linewidth]{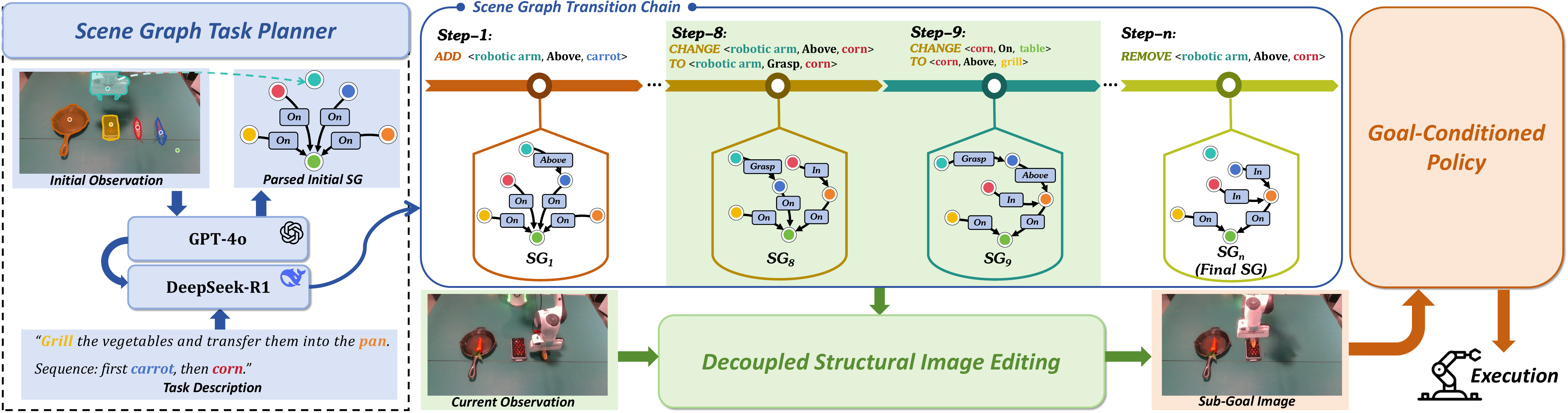}
    \caption{An overview of our proposed framework SAGE. The scene graph task planner first parses the scene state from the initial observation, then converts the task description into a scene graph transition chain. For each transition step in this chain, the decoupled structural image editing pipeline translates the target sub-goal scene graph into an RGB image. Conditioned on this sub-goal image, our goal-conditioned policy executes actions to reach the desired state. SAGE accomplishes an entire long-horizon manipulation task by iteratively switching goals and executing actions.}
    \label{fig:overview}
\end{figure*}

\section{Preliminaries}
\smallskip
\noindent\textbf{Scene Graph.} 
We use a semantic scene graph \cite{semantic_chang2021comprehensive} to represent the complex scene state, denoted as $\mathcal{G} = \{\mathcal{V}, \mathcal{E}\}$. Given $N_o$ objects, $\mathcal{V} = \{v_i \mid i=1,2,...,N_o\}$ refers to the set of object nodes. $\mathcal{E} \subseteq \mathcal{V} \times \mathcal{R} \times \mathcal{V}$ is the set of directed edges, where each edge $e_{ij} = (v_i,r,v_j)$ represents a relationship $r$ from object $v_i$ to object $v_j$. In this paper, $\mathcal{R}$ contains the following relationships: \{\texttt{Above}, \texttt{On}, \texttt{In}, \texttt{Grasp}, \texttt{Next To}\}.

\smallskip
\noindent\textbf{Scene Graph Transition Chain.}
For a multi-object, long-horizon manipulation task, the process is modeled as a sequence of scene state transitions. We represent such transitions as a scene graph transition chain $\mathcal{C}=\{\mathcal{G}_0, \mathcal{G}_1, ..., \mathcal{G}_N\}$, where $N$ is the number of transition steps and $\mathcal{G}_k \to \mathcal{G}_{k+1}$ represents a single transition step from the current scene state to the next.

\section{Method}
Fig. \ref{fig:overview} presents an overview of our framework SAGE. Given an RGB observation and a task description, our scene graph task planner converts the task into a scene graph transition chain (Section.\ref{planning}). For each transition step in this chain, our decoupled structural image editing pipeline translates the target sub-goal scene graph into a corresponding RGB image (Section.\ref{image_editing}). This sub-goal image is then provided to our goal-conditioned policy, which executes actions to achieve the desired state transition (Section.\ref{policy}). SAGE completes an entire long-horizon manipulation task by iteratively switching goals and executing actions.

\subsection{Scene Graph Task Planner}
\label{planning}
The scene graph task planner intergrates a VLM (i.e. GPT-4o) and a LLM (i.e. DeepSeek-R1) to convert a high-level task description into a physically-grounded symbolic plan, as illustrated in the left side of Fig. \ref{fig:overview}. 

The planner begins by employing the VLM to parse the initial scene graph $\mathcal{G}_0$ from the input initial observation $O_0$ via Chain-of-Thought (CoT) \cite{cot_wei2022chain}. We adopt a hybrid reasoning approach that combines both geometric and semantic information to ensure the correctness of the results. Specifically, object identities and their corresponding bounding boxes are first provided to the VLM. The VLM then predicts pairwise spatial relationships by selecting from a predefined set of symbolic relationships (i.e.  \{\texttt{Above}, \texttt{On}, \texttt{In}, \texttt{Grasp}, \texttt{Next To}\}) based on both visual context and the geometric information, thereby constructing the initial scene graph $\mathcal{G}_0$.


The constructed scene graph $\mathcal{G}_0$ and the task description $\tau$ are provided to the LLM, which performs spatial and causal reasoning to infer a scene graph transition chain $\mathcal{C}$.  We impose a transition constraint that each step $\mathcal{G}_k \rightarrow \mathcal{G}_{k+1}$ must change exactly one relationship edge intentionally, representing a discrete high-level robot action such as lifting and releasing. This restriction serves two purposes: (1) it simplifies the spatial reasoning process for the LLM by structuring the output space into tractable steps, and (2) it guarantees that each transition corresponds to a new, physically-grounded state in the task's progression. To further enhance planning robustness, we embed other structural constraints into the LLM's reasoning process, such as relationship directionality and action prerequisites (e.g., \texttt{Grasp} must be preceded by \texttt{Above}). This allows the LLM to generate a coherent and executable symbolic plan for the entire task.


\begin{figure*}
    \centering    
    \includegraphics[width=\linewidth]{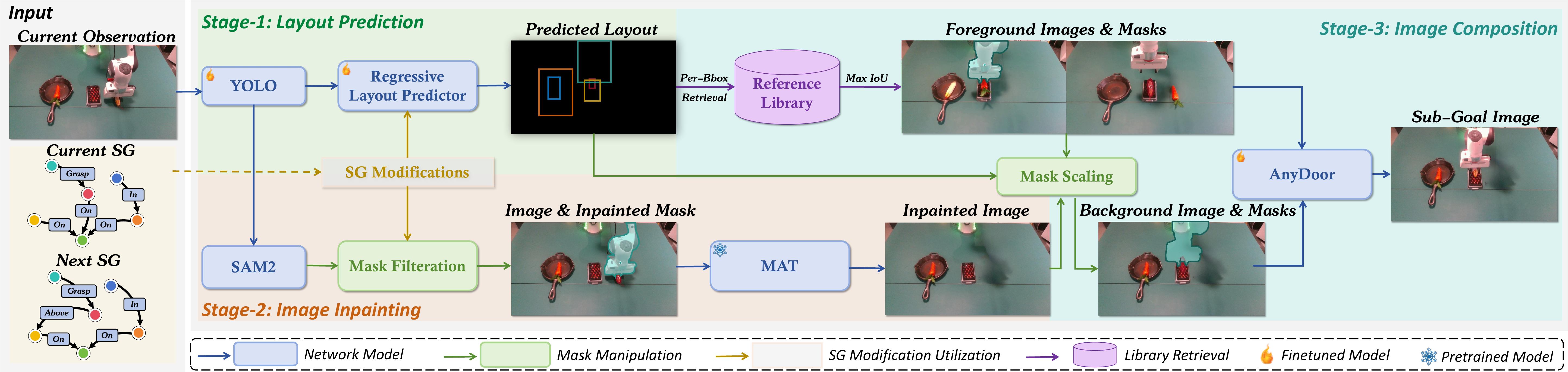}
    \caption{The workflow of our decoupled structural image editing pipeline. In stage 1, we leverage regressive layout predictors to predict layout corresponding to the next scene graph. In stage 2, we use SAM2 \cite{sam2_ravi2024sam} and MAT \cite{mat_li2022mat} to inpaint the current observation and remove objects. In stage 3, based on a pre-built reference demonstration library, we utilize AnyDoor \cite{anydoor_chen2024anydoor} to compose foreground and background images, inserting objects into target positions.}
    \label{fig:editing_pipeline}
\end{figure*}

\subsection{Decoupled Structural Image Editing}
\label{image_editing}

Inspired by \cite{sgedit_zhang2024sgedit}, the transitions between scene graphs can be viewed as object movements. We thus design a direct and effective way to realize these movements, i.e., first erasing the object from its original location and then inserting it into a new one. To this end, as shown in Fig. \ref{fig:editing_pipeline}, we propose a structural image editing pipeline decoupled into three stages: layout prediction, image inpainting and image composition. The pipeline significantly outperforms other end-to-end text-to-image generation models and demonstrates superior generation controllability (detailed in Section. \ref{image_editing_experiments}).

\smallskip
\noindent\textbf{Layout Prediction.} 
For any scene graph transition $\mathcal{G}_k \to \mathcal{G}_{k+1}$ in chain $\mathcal{C}$,  we first extract the object layout $L_k$ from current observation $O_k$. While open-vocabulary methods like GroundingDino \cite{liu2024grounding} can also be applied in this operation, we empirically found that they have difficulty in stably detecting the robotic arm or similar objects with different colors. We therefore train YOLO \cite{yolo_redmon2016you} for this extraction. 

We then predict the next layout $L_{k+1}$ by:
\begin{equation}
    L_{k+1} = \mathcal{P}(\mathcal{G}_k,L_k,\mathcal{G}_{k+1}), \quad k \in \{1,2,...,N-1\},
\end{equation}
where $\mathcal{P}$ is the layout predictor, $N$ is the length of the scene graph transition chain $\mathcal{C}$. This process ensures that our image editing pipeline has the necessary geometric information to place objects correctly.

The predictor $\mathcal{P}$ is formed by a set of linear regression models for bounding box prediction. While more complex models, such as transformers \cite{vaswani2017attention}, could be used for more complex scenarios such as camera viewpoint changes or occlusion, linear regression is sufficient for our tasks.




\smallskip
\noindent\textbf{Image Inpainting.} 
Using current object layout $L_k$ as a guide, we first use SAM2 \cite{sam2_ravi2024sam} to segment current observed image $I_k$,  producing a set of object masks $\mathcal{M}_k$. 

We then identify a set of target objects, $\mathcal{V}_f \subseteq \mathcal{V}$, whose relationships have changed in the scene graph transition $\mathcal{G}_k \to \mathcal{G}_{k+1}$. We filter $\mathcal{M}_k$ to keep only masks of these target objects, creating a final set of masks denoted a $\mathcal{M}_k^f$.

Finally, we utilize a pretrained MAT \cite{mat_li2022mat} model to remove these objects defined by $\mathcal{M}_k^f$ from $I_k$, generating an inpainted background image $I_{bg}$:
\begin{equation}
    I_{bg} = \mathcal{F}(I_k, \mathcal{M}_k^f)
\end{equation}
where $\mathcal{F}$ represents the inpainting function.

We have also tried shadow removal \cite{hu2024unveiling} in this stage, but found that it had no discernible impact on the performance of the sub-goal images. For simplicity, this step was excluded.

\smallskip
\noindent\textbf{Image Composition.}
Using the predicted layout $L_{k+1}$ and the inpainted background $I_{bg}$, we compose the sub-goal image $I_{k+1}$ as follows.

For each object $v_i \in \mathcal{V}_f$, we retrieve the optimal reference $v_i^*$ from a pre-built demonstration library $\mathcal{L}$ containing sampled images, layouts, and masks:
\begin{equation}
    v_i^* = \arg \max \limits_{v_{i^\prime} \in \mathcal{L}_i}  
    \begin{cases}
    \text{IoU}(B_{i}, B_{i^\prime}) & \text{if IoU} > 0 \\
    -\text{dist}(c_{i}, c_{i^\prime}) & \text{if IoU} = 0 \\
    \end{cases}
\end{equation}
where $\mathcal{L}_i \subset \mathcal{L}$ denotes the subset of objects with the same label as $v_i$, $B_{i}$ and $c_{i}$ are the predicted bounding box and center for $v_i$ in layout $L_{k+1}$, $B_{i'}$ and $c_{i'}$ are those for a candidate object $v_{i'}$ from the library. We then collect the corresponding optimal images, masks, and boxes into sets $\mathcal{I}_{fg}$, $\mathcal{M}_{fg}$, and $\mathcal{B}_{fg}$.

Next, we generate the background masks $\mathcal{M}_{bg}$ for the inpainted image by scaling and repositioning the retrieved foreground masks to align with the predicted layout:
\begin{equation}
    \mathcal{M}_{bg} = G(\mathcal{M}_{fg}, \mathcal{B}_{fg}, L_{k+1})
\end{equation} 
where $G$ is the mask generation function.

The sub-goal image $I_{k+1}$ is thus synthesized by compositing the foreground objects onto $I_{bg}$ using the masks $\mathcal{M}_{fg}$ and $\mathcal{M}_{bg}$. Specifically, we finetune AnyDoor \cite{anydoor_chen2024anydoor} to ensure physical and kinematic feasibility for the robotic arm:
\begin{equation}
    I_{k+1} = \mathcal{H}(\mathcal{I}_{fg}, I_{bg},\mathcal{M}_{fg},\mathcal{M}_{bg})
\end{equation}
where $\mathcal{H}$ represents the image composition function.

\subsection{Goal-Conditioned Policy} 
\label{policy}
Our goal-conditioned policy is built upon the ACT \cite{act_zhao2023learning} architecture, with an additional image backbone for sub-goal image processing. At each transition step $\mathcal{G}_k \to \mathcal{G}_{k+1}$, conditioned on the sub-goal image $I_{k+1}$, the policy continuously takes in observations and generates actions to reach the desired sub-goal state. 

\smallskip
\noindent\textbf{Training.} 
We largely follow ACT \cite{act_zhao2023learning} to train our policy, but incorporate additional goal images in the training data. Specifically, we employ UVD \cite{uvd_zhang2024universal} to decompose the demonstration videos into segments, using the final frame of each segment as a sub-goal image for that phase.

\smallskip
\noindent\textbf{Inference.} 
At inference time, the policy acquires a sub-goal image depending on the task type: (1) For unseen tasks, the sub-goal image is provided by our image editing pipeline; (2) For seen tasks, a corresponding real image is retrieved directly from the reference library.

Inspired by MILES \cite{miles_papagiannis2024miles},  we use a FIFO buffer to store predicted actions. When the action difference within the buffer falls below a threshold $\delta$, we consider the sub-goal achieved and the transition step should proceed forward.

\begin{figure*}
    \centering    
    \includegraphics[width=\linewidth]{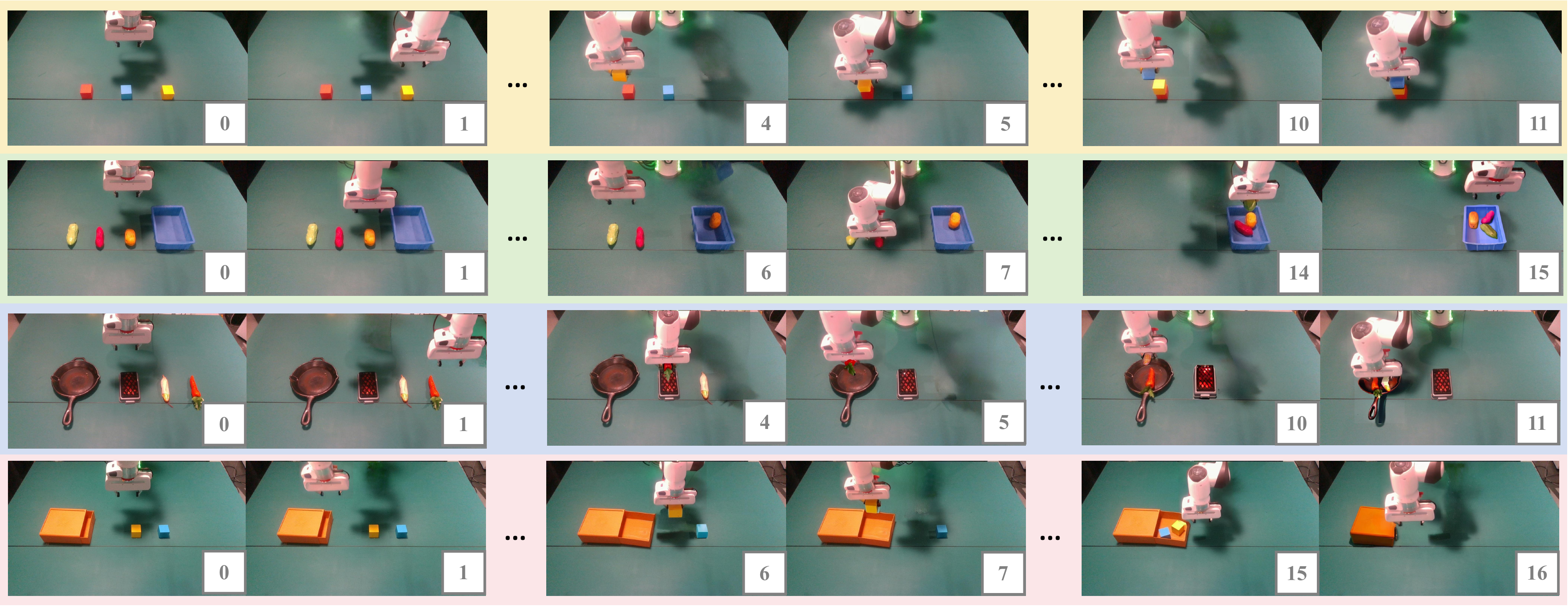}
    \caption{Images generated by our decoupled structural image editing pipeline, showing the robot's state transitions for different long-horizon tasks. From top to down: Sequential Task, Flexible Task, Hybrid Task \uppercase\expandafter{\romannumeral1}, Hybrid Task \uppercase\expandafter{\romannumeral2}.}
    \label{fig:rollouts}
\end{figure*}

\begin{table*}[h]
\caption{Task Execution Comparison - \textbf{Task Success Rate (Phase Success Rate)}}
\label{table1}
\begin{center}

\resizebox{\linewidth}{!}{
\begin{tabular}{c c c c c c c c c}
\toprule
\multirow{2}{*}{Method} & \multicolumn{2}{c}{Sequential Task} & \multicolumn{2}{c}{Flexible Task} & \multicolumn{2}{c}{Hybrid Task \uppercase\expandafter{\romannumeral1} } & \multicolumn{2}{c}{Hybrid Task \uppercase\expandafter{\romannumeral2}} \\
\cmidrule(lr){2-3} \cmidrule(lr){4-5} \cmidrule(lr){6-7} \cmidrule(lr){8-9}
& Seen & Unseen & Seen & Unseen & Seen & Unseen & Seen & Unseen\\
\midrule
Pi-0 \cite{pi0_black2024pi_0} & 2/10 (44.1\%) & 0/30 (19.7\%) & 1/10 (37.9\%) & 0/30 (20.1\%) & 3/10 (57.8\%) & 0/30 (31.7\%) & 2/10 (44.8\%) & 0/30 (32.5\%)\\
OpenVLA \cite{openvla_kim2024openvla} & 1/10 (36.2\%) & 0/30 (17.1\%) & 0/10 (22.1\%) & 0/30 (9.6\%) & 1/10 (34.0\%) & 0/30 (20.7\%) & 2/10 (41.1\%) & 0/30 (29.7\%)\\
ACT \cite{act_zhao2023learning} & 0/10 (24.6\%) & - & 0/10 (21.6\%) & - & 0/10 (17.0\%) & -  &  0/10 (16.3\%) &  -\\
Diffusion Policy \cite{dp_chi2023diffusion} & 0/10 (17.1\%) & - & 0/10 (17.6\%) & - & 0/10 (14.8\%) & - &  0/10 (9.4\%) &  -\\
\midrule
\textbf{Our Method} & \textbf{10/10 (100.0\%)} & \textbf{25/30 (91.3\%)} & \textbf{10/10 (100.0\%)} & \textbf{20/30 (73.2\%)} & \textbf{10/10 (100.0\%)} & \textbf{26/30 (85.0\%)}  & \textbf{10/10 (100.0\%)} & \textbf{22/30 (81.3\%)} \\
\bottomrule
\end{tabular}}
\end{center}
\end{table*}

\section{Experiments}
In this section, we conduct a series of experiments to evaluate the effectiveness of our proposed framework SAGE and answer the following four key questions:

\textbf{Q1:} Does our framework SAGE effectively align task-level symbolic planning and pixel-level visuo-motor control?

\textbf{Q2:} Can our goal-conditioned policy, which completes long-horizon tasks phase by phase, outperforms single-task policies that tackle the entire task at once?

\textbf{Q3:} How robust is our scene graph task planner in generating physically-grounded and executable scene graph transition chains?

\textbf{Q4:} Do edited images faithfully represent sub-goal states and guide policy in unseen tasks?

\subsection{Implementation Details}

\smallskip
\noindent\textbf{Tasks and Dataset.}
We consider three different task classes in our experiments:
\begin{itemize}
    \item \textbf{Sequential Task.} This task class requires a strict execution order. We use a block stacking task with three different colored blocks, where the robot must stack them in a specific color sequence. 

    \item \textbf{Flexible Task.} This task class allows flexible ordering of execution steps. We use a fruit placement task, where three different fruits must be placed in a box. 

    \item \textbf{Hybrid Task.} This task class integrates both flexible and sequential operations. Two tasks are employed: (1) a barbecue task, where vegetables can be chosen flexibly, but must then be sequentially grilled and transferred to a pan; (2) a drawer storage task, where the drawer is first pulled out, blocks are then selected and placed inside flexibly, and the drawer is finally pushed back.
    
\end{itemize}
We collected 30 demonstrations per task, varying the initial object placements while keeping the execution sequence the same. To bridge the domain gap, the training dataset incorporates edited images by applying in-place editing to real sub-goal images. We define a \textbf{seen task} as one that follows the execution sequence found in the training data, while an \textbf{unseen task} as any task with a different sequence.


\smallskip
\noindent\textbf{Hardware Setup.} 
We utilize two Intel RealSense D435i RGB cameras with 1280x720 resolution. One camera captures a third-person view of the scene, while the other is mounted on the 7-DoF Franka Emika Panda robotic arm to provide an arm view. Note that for planning and image editing, only the third-person camera view is used. The policy receives observations from both the third-person view and arm view during execution.

\smallskip
\noindent\textbf{Training Setup.} 
The goal-conditioned policy is trained for 100 epochs with a batch size of 8 and a chunk size of 100. The YOLOv11 model is trained for 50 epochs on a dataset of 300 labeled frames each task, using a batch size of 16. AnyDoor finetuning uses 8000 frames randomly sampled from the demonstration videos, with a batch size of 16 for 2 epochs. All models were trained on two Nvidia A100 SXM4 80GB GPUs.

\begin{figure*}
    \centering
    \includegraphics[width=\linewidth]{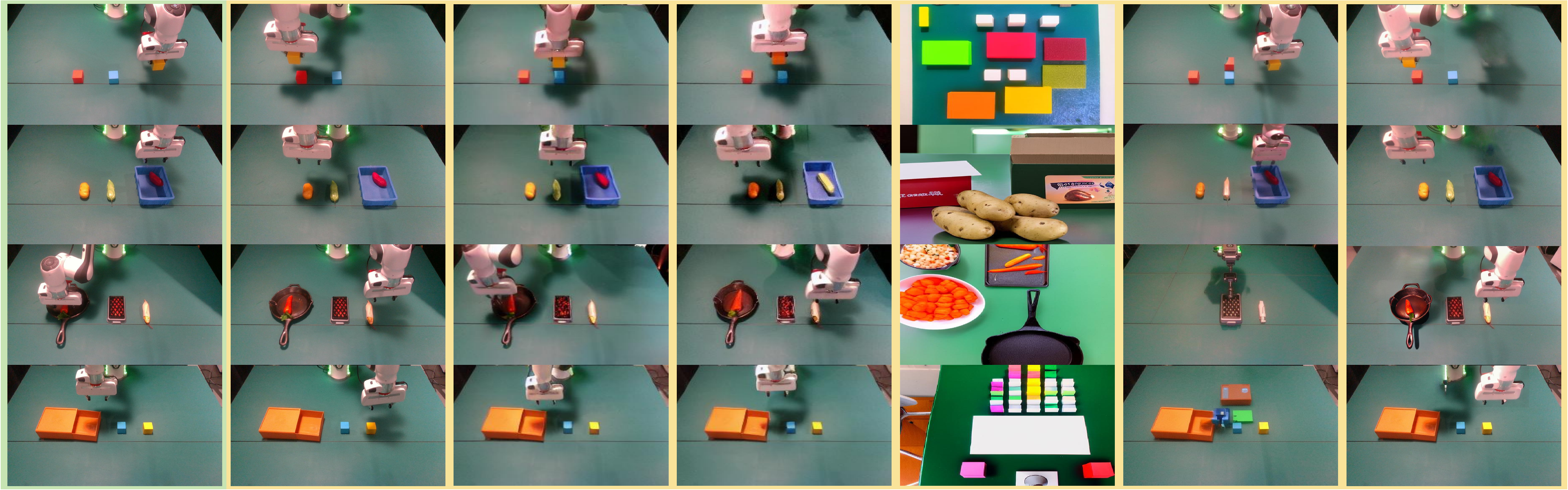}
    \parbox[t]{1.0\linewidth}{\relax
                           \hspace{28px} (a) \hspace{53px} 
                           (b) \hspace{53px} (c) \hspace{53px} (d) \hspace{55px} (e) \hspace{55px} (f) \hspace{55px} (g)}
    \caption{Qualitative comparison with other baseline methods. From left to right: (a) Current observation. (b) Ground-truth next sub-goal image. (c) SuSIE \cite{susie_black2023zero}. (d) GR-MG \cite{grmg_li2025gr}. (e) Break-A-Scene \cite{break-a-scene_avrahami2023break}. (f) SGEdit \cite{sgedit_zhang2024sgedit}. (g) Our decoupled structural image editing pipeline. }
    \label{fig:editing_comparison}
\end{figure*}

\begin{table*}[h]
\caption{Image Editing Performance Evaluation on - User Score, LPIPS, Task Success Rate (Phase Success Rate)}
\label{table2}
\begin{center}
\resizebox{\linewidth}{!}{
\begin{tabular}{c c c c c c c}
\toprule
\multirow{2}{*}{Method} & \multirow{2}{*}{Score ($\uparrow$)} & \multirow{2}{*}{LPIPS ($\downarrow$)} & \multicolumn{4}{c}{Task Success Rate (Phase Success Rate) ($\uparrow$)} \\
\cmidrule{4-7}
& & & Unseen Sequential Task & Unseen Flexible Task & Unseen Hybrid Task \textrm{I} &  Unseen Hybrid Task \textrm{II} \\
\midrule
SuSIE \cite{susie_black2023zero} & 4.8/10 & 0.55 & 0/30 (0.0\%) & 0/30 (12.9\%) & 0/30 (34.8\%) & 0/30 (23.1\%) \\
GR-MG \cite{grmg_li2025gr} & 6.4/10 & 0.49 & 0/30 (0.0\%) & 0/30 (11.1\%) & 0/30 (21.2\%) & 0/30 (18.4\%) \\
Break-A-Scene \cite{break-a-scene_avrahami2023break} & 1.2/10 & 0.88 & 0/30 (0.0\%) & 0/30 (0.0\%) & 0/30 (0.0\%) & 0/30 (0.0\%)  \\
SGEdit \cite{sgedit_zhang2024sgedit} & 3.0/10 & 0.47 & 0/30 (0.0\%) & 0/30 (0.0\%) & 0/30 (0.0\%) & 0/30 (0.0\%) \\
\midrule
\textbf{Our Method} & \textbf{9.4/10} & \textbf{0.26} & \textbf{25/30 (91.3\%)} & \textbf{20/30 (73.2\%)} & \textbf{26/30 (85.0\%)} & \textbf{22/30 (81.3\%)} \\
\bottomrule
\end{tabular} 
}
\end{center}
\end{table*}

\subsection{Evaluation Protocols}

\smallskip

We compare our approach against two categories of baselines.

\smallskip
\noindent \textbf{End-to-End Execution.} We benchmark our approach against VLM-based planners and single-task policies: (1) Pi-0 \cite{pi0_black2024pi_0} and OpenVLA \cite{openvla_kim2024openvla}, which translate high-level commands into action sequences. We evaluate them on both seen and unseen tasks with random temporal execution orderings. (2) ACT \cite{act_zhao2023learning} and Diffusion Policy \cite{dp_chi2023diffusion}, which are designed for single-task execution. Since they cannot generalize to unseen tasks, we only assess their performance on seen tasks.

\smallskip
\noindent \textbf{Sub-Goal Image Generation.} We compare our decoupled structural image editing pipeline with image editing methods: (1) SuSIE \cite{susie_black2023zero} and GR-MG \cite{grmg_li2025gr} , which are IP2P \cite{ip2p_brooks2023instructpix2pix}-based editing models pretrained on large-scale robotic datasets. We extend their prediction window to match our scene graph transition interval. (2) Break-A-Scene \cite{break-a-scene_avrahami2023break} and SGEdit \cite{sgedit_zhang2024sgedit}, which interpret semantic concepts in an image and generate new ones from text instructions. To ensure compatibility, we convert each target scene graph into textual descriptions using an LLM before providing them as input.

\smallskip
For a fair comparison, all training-based baselines are trained or finetuned on our dataset. When evaluating other image editing methods, the corresponding policies are trained by incorporating newly edited images from these methods.

\smallskip
\noindent\textbf{Metrics.}
We measure the task performance in two metrics: (1) Task Success Rate, which measures whether the entire task is completed; (2) Phase Success Rate, which measures the proportion of successfully completed phases relative to the total number of phases in the decomposed tasks.

To measure how accurately the generated images represent the intended sub-goal states, we use (1) LPIPS \cite{lpips_zhang2018unreasonable} to quantify perceptual similarity between the generated and the ground-truth images. We also adopt (2) User Score, where 10 human evaluators rate each image from 0 to 10, based on visual quality and semantic alignment between the generated and the ground-truth sub-goal state. The final result is calculated as an average.

\subsection{Experiment Results in End-to-End Execution}
As the experimental results shown in Table. \ref{table1}, we answer three questions:

\smallskip
\noindent\textbf{Answer to Q1.} The results demonstrate that our framework significantly outperforms other baselines, achieving strong performance across both seen and unseen variations of all long-horizon tasks. While models like Pi-0 \cite{pi0_black2024pi_0} and OpenVLA \cite{openvla_kim2024openvla} are capable of performing implicit, language-based planning, they fail to successfully convert the plan into low-level control and stably complete task phases. This indicates the inherent difficulty of aligning unstructured language plans with continuous control. In contrast, our approach provides a physically-grounded plan via the scene graph, enabling our method to robustly handle the complexities of long-horizon manipulation.


\smallskip
\noindent\textbf{Answer to Q2.} Our goal-conditioned policy, which completes long-horizon tasks phase by phase, significantly outperforms single-task policies. While ACT \cite{act_zhao2023learning} and Diffusion Policy \cite{dp_chi2023diffusion} can complete initial task phases, they fail to finish the entire long-horizon tasks. In contrast, our goal-conditioned policy successfully completes various phases using different sub-goal images. This indicates that rather than trying to solve the entire long-horizon task at once, it is more effective to learn to achieve multiple sub-goals. It also confirms that our policy dynamically reacts to sub-goal images to infer actions, which is a key capability for generalizing to unseen scenarios.

\smallskip
\noindent\textbf{Answer to Q3.} The high task success rates of SAGE (Table. \ref{table1}) are supported by the robustness of our scene graph task planner. Out of the 160 execution trials across all tasks, only 12 produced partially-infeasible transition chains. The VLM for initial scene graph parsing exhibited high accuracy in identifying objects and inferring their spatial relationships. The rare failures were attributed to the LLM-based reasoning, which occasionally introduced logical errors such as redundant action loops. For example, in the flexible task, after correctly planning to place a fruit into the box, the LLM sometimes generated subsequent transitions to re-grasp and re-place the same fruit. 
Exemplar scene graph transition chains are provided in the supplementary video, demonstrating the scene graph task planner’s effectiveness in producing physically-grounded task plans.


\subsection{Experiment Results in Sub-Goal Image Generation}
\label{image_editing_experiments}
We summarize quantitative results in Table.~\ref{table2}, and illustrate qualitative examples in Fig. \ref{fig:rollouts} and Fig. \ref{fig:editing_comparison}. Together, these results answer the final key question:

\noindent\textbf{Answer to Q4.}
Our decoupled structural image editing pipeline produces high-quality sub-goal images that accurately represent desired states, achieving a user score of 9.4 and an LPIPS value of 0.26. In contrast, while SuSIE \cite{susie_black2023zero} and GR-MG \cite{grmg_li2025gr} generate harmonized images, they often fail to correctly interpret unseen task descriptions, resulting in semantically incorrect sub-goal representations (e.g., the robotic arm is in the wrong position). The challenge is even greater for Break-A-Scene \cite{break-a-scene_avrahami2023break} and SGEdit \cite{sgedit_zhang2024sgedit}, which fail to understand the robot manipulation scene, generating corrupted images without meaningful guidance. The high accuracy of our generated sub-goal images is what enables our policy's high phase success rate of 82.7\% on average.

\section{Conclusions \& Future Work}

In this paper, we presented a novel framework \textbf{SAGE} for long-horizon manipulation that utilizes structural semantic scene graphs to bridge task-level semantic reasoning and pixel-level visuo-motor control. Our framework consists of a scene graph task planner and a decoupled structural image editing pipeline. The planner translates the task into a scene graph transition chain, which is then converted into a sequence of sub-goal images. These images are used to guide a visuo-motor policy, enabling the robot to accomplish the entire long-horizon task by iteratively completing sub-goals. Experimental results demonstrate the superiority of our framework in solving long-horizon manipulation tasks.

There are a few promising directions to further explore. First, incorporating 3D geometric and contact information could enable SAGE to extend to more cluttered environments. Second, rather than open-loop, error detection and recovery would allow closed-loop planning in the task level. Finally, advances in open-vocabulary perception will enable broader generalization to novel objects, further expanding the applicability of the framework.








\bibliographystyle{IEEEtran}
\bibliography{literature}

\begin{thebibliography}{10}
\providecommand{\url}[1]{#1}
\csname url@samestyle\endcsname
\providecommand{\newblock}{\relax}
\providecommand{\bibinfo}[2]{#2}
\providecommand{\BIBentrySTDinterwordspacing}{\spaceskip=0pt\relax}
\providecommand{\BIBentryALTinterwordstretchfactor}{4}
\providecommand{\BIBentryALTinterwordspacing}{\spaceskip=\fontdimen2\font plus
\BIBentryALTinterwordstretchfactor\fontdimen3\font minus \fontdimen4\font\relax}
\providecommand{\BIBforeignlanguage}[2]{{%
\expandafter\ifx\csname l@#1\endcsname\relax
\typeout{** WARNING: IEEEtran.bst: No hyphenation pattern has been}%
\typeout{** loaded for the language `#1'. Using the pattern for}%
\typeout{** the default language instead.}%
\else
\language=\csname l@#1\endcsname
\fi
#2}}
\providecommand{\BIBdecl}{\relax}
\BIBdecl

\bibitem{feng2025reflective}
Y.~Feng, J.~Han, Z.~Yang, X.~Yue, S.~Levine, and J.~Luo, ``Reflective planning: Vision-language models for multi-stage long-horizon robotic manipulation,'' \emph{arXiv preprint arXiv:2502.16707}, 2025.

\bibitem{luo2024multistage}
J.~Luo, C.~Xu, X.~Geng, G.~Feng, K.~Fang, L.~Tan, S.~Schaal, and S.~Levine, ``Multistage cable routing through hierarchical imitation learning,'' \emph{IEEE Transactions on Robotics}, vol.~40, pp. 1476--1491, 2024.

\bibitem{kappler2018real}
D.~Kappler, F.~Meier, J.~Issac, J.~Mainprice, C.~G. Cifuentes, M.~W{\"u}thrich, V.~Berenz, S.~Schaal, N.~Ratliff, and J.~Bohg, ``Real-time perception meets reactive motion generation,'' \emph{IEEE Robotics and Automation Letters}, vol.~3, no.~3, pp. 1864--1871, 2018.

\bibitem{murphy2019introduction}
R.~R. Murphy, \emph{Introduction to AI robotics}.\hskip 1em plus 0.5em minus 0.4em\relax MIT press, 2019.

\bibitem{garrett2021integrated}
C.~R. Garrett, R.~Chitnis, R.~Holladay, B.~Kim, T.~Silver, L.~P. Kaelbling, and T.~Lozano-P{\'e}rez, ``Integrated task and motion planning,'' \emph{Annual review of control, robotics, and autonomous systems}, vol.~4, no.~1, pp. 265--293, 2021.

\bibitem{kaelbling2011hierarchical}
L.~P. Kaelbling and T.~Lozano-P{\'e}rez, ``Hierarchical task and motion planning in the now,'' in \emph{2011 IEEE international conference on robotics and automation}.\hskip 1em plus 0.5em minus 0.4em\relax IEEE, 2011, pp. 1470--1477.

\bibitem{huang2022language}
W.~Huang, P.~Abbeel, D.~Pathak, and I.~Mordatch, ``Language models as zero-shot planners: Extracting actionable knowledge for embodied agents,'' in \emph{International conference on machine learning}.\hskip 1em plus 0.5em minus 0.4em\relax PMLR, 2022, pp. 9118--9147.

\bibitem{huang2022inner}
W.~Huang, F.~Xia, T.~Xiao, H.~Chan, J.~Liang, P.~Florence, A.~Zeng, J.~Tompson, I.~Mordatch, Y.~Chebotar \emph{et~al.}, ``Inner monologue: Embodied reasoning through planning with language models,'' \emph{arXiv preprint arXiv:2207.05608}, 2022.

\bibitem{ahn2022can}
M.~Ahn, A.~Brohan, N.~Brown, Y.~Chebotar, O.~Cortes, B.~David, C.~Finn, C.~Fu, K.~Gopalakrishnan, K.~Hausman \emph{et~al.}, ``Do as i can, not as i say: Grounding language in robotic affordances,'' \emph{arXiv preprint arXiv:2204.01691}, 2022.

\bibitem{rana2023sayplan}
K.~Rana, J.~Haviland, S.~Garg, J.~Abou-Chakra, I.~Reid, and N.~Suenderhauf, ``Sayplan: Grounding large language models using 3d scene graphs for scalable robot task planning,'' \emph{arXiv preprint arXiv:2307.06135}, 2023.

\bibitem{driess2023palm}
D.~Driess, F.~Xia, M.~S. Sajjadi, C.~Lynch, A.~Chowdhery, A.~Wahid, J.~Tompson, Q.~Vuong, T.~Yu, W.~Huang \emph{et~al.}, ``Palm-e: An embodied multimodal language model,'' 2023.

\bibitem{rt1_brohan2022rt}
A.~Brohan, N.~Brown, J.~Carbajal, Y.~Chebotar, J.~Dabis, C.~Finn, K.~Gopalakrishnan, K.~Hausman, A.~Herzog, J.~Hsu \emph{et~al.}, ``Rt-1: Robotics transformer for real-world control at scale,'' \emph{arXiv preprint arXiv:2212.06817}, 2022.

\bibitem{huang2023voxposer}
W.~Huang, C.~Wang, R.~Zhang, Y.~Li, J.~Wu, and L.~Fei-Fei, ``Voxposer: Composable 3d value maps for robotic manipulation with language models,'' \emph{arXiv preprint arXiv:2307.05973}, 2023.

\bibitem{shi2024yell}
L.~X. Shi, Z.~Hu, T.~Z. Zhao, A.~Sharma, K.~Pertsch, J.~Luo, S.~Levine, and C.~Finn, ``Yell at your robot: Improving on-the-fly from language corrections,'' \emph{arXiv preprint arXiv:2403.12910}, 2024.

\bibitem{rt2_zitkovich2023rt}
B.~Zitkovich, T.~Yu, S.~Xu, P.~Xu, T.~Xiao, F.~Xia, J.~Wu, P.~Wohlhart, S.~Welker, A.~Wahid \emph{et~al.}, ``Rt-2: Vision-language-action models transfer web knowledge to robotic control,'' in \emph{Conference on Robot Learning}.\hskip 1em plus 0.5em minus 0.4em\relax PMLR, 2023, pp. 2165--2183.

\bibitem{pi0_black2024pi_0}
K.~Black, N.~Brown, D.~Driess, A.~Esmail, M.~Equi, C.~Finn, N.~Fusai, L.~Groom, K.~Hausman, B.~Ichter \emph{et~al.}, ``pi\_0: A vision-language-action flow model for general robot control,'' \emph{arXiv preprint arXiv:2410.24164}, 2024.

\bibitem{openvla_kim2024openvla}
M.~J. Kim, K.~Pertsch, S.~Karamcheti, T.~Xiao, A.~Balakrishna, S.~Nair, R.~Rafailov, E.~Foster, G.~Lam, P.~Sanketi \emph{et~al.}, ``Openvla: An open-source vision-language-action model,'' \emph{arXiv preprint arXiv:2406.09246}, 2024.

\bibitem{susie_black2023zero}
K.~Black, M.~Nakamoto, P.~Atreya, H.~Walke, C.~Finn, A.~Kumar, and S.~Levine, ``Zero-shot robotic manipulation with pretrained image-editing diffusion models,'' \emph{arXiv preprint arXiv:2310.10639}, 2023.

\bibitem{grmg_li2025gr}
P.~Li, H.~Wu, Y.~Huang, C.~Cheang, L.~Wang, and T.~Kong, ``Gr-mg: Leveraging partially-annotated data via multi-modal goal-conditioned policy,'' \emph{IEEE Robotics and Automation Letters}, 2025.

\bibitem{ho2020denoising}
J.~Ho, A.~Jain, and P.~Abbeel, ``Denoising diffusion probabilistic models,'' \emph{Advances in neural information processing systems}, vol.~33, pp. 6840--6851, 2020.

\bibitem{ip2p_brooks2023instructpix2pix}
T.~Brooks, A.~Holynski, and A.~A. Efros, ``Instructpix2pix: Learning to follow image editing instructions,'' in \emph{Proceedings of the IEEE/CVF conference on computer vision and pattern recognition}, 2023, pp. 18\,392--18\,402.

\bibitem{semantic_chang2021comprehensive}
X.~Chang, P.~Ren, P.~Xu, Z.~Li, X.~Chen, and A.~Hauptmann, ``A comprehensive survey of scene graphs: Generation and application,'' \emph{IEEE Transactions on Pattern Analysis and Machine Intelligence}, vol.~45, no.~1, pp. 1--26, 2021.

\bibitem{paul1981robot}
R.~P. Paul, \emph{Robot manipulators: mathematics, programming, and control: the computer control of robot manipulators}.\hskip 1em plus 0.5em minus 0.4em\relax Richard Paul, 1981.

\bibitem{whitney1972mathematics}
D.~E. Whitney, ``The mathematics of coordinated control of prosthetic arms and manipulators,'' 1972.

\bibitem{vukobratovic1984scientific}
M.~Vukobratovic, V.~Potkonjak, D.~Stokic, and B.~Roth, ``Scientific fundamentals of robotics 1: Dynamics of manipulation robots, theory and application and scientific fundamentals of robotics 2: Control of manipulation, robots, theory and application,'' 1984.

\bibitem{dantam2016incremental}
N.~T. Dantam, Z.~K. Kingston, S.~Chaudhuri, and L.~E. Kavraki, ``Incremental task and motion planning: A constraint-based approach.'' in \emph{Robotics: Science and systems}, vol.~12.\hskip 1em plus 0.5em minus 0.4em\relax Ann Arbor, MI, USA, 2016, p. 00052.

\bibitem{migimatsu2020object}
T.~Migimatsu and J.~Bohg, ``Object-centric task and motion planning in dynamic environments,'' \emph{IEEE Robotics and Automation Letters}, vol.~5, no.~2, pp. 844--851, 2020.

\bibitem{srivastava2014combined}
S.~Srivastava, E.~Fang, L.~Riano, R.~Chitnis, S.~Russell, and P.~Abbeel, ``Combined task and motion planning through an extensible planner-independent interface layer,'' in \emph{2014 IEEE international conference on robotics and automation (ICRA)}.\hskip 1em plus 0.5em minus 0.4em\relax IEEE, 2014, pp. 639--646.

\bibitem{wang2021learning}
Z.~Wang, C.~R. Garrett, L.~P. Kaelbling, and T.~Lozano-P{\'e}rez, ``Learning compositional models of robot skills for task and motion planning,'' \emph{The International Journal of Robotics Research}, vol.~40, no. 6-7, pp. 866--894, 2021.

\bibitem{driess2020deep}
D.~Driess, J.-S. Ha, and M.~Toussaint, ``Deep visual reasoning: Learning to predict action sequences for task and motion planning from an initial scene image,'' \emph{arXiv preprint arXiv:2006.05398}, 2020.

\bibitem{chane2021goal}
E.~Chane-Sane, C.~Schmid, and I.~Laptev, ``Goal-conditioned reinforcement learning with imagined subgoals,'' in \emph{International conference on machine learning}.\hskip 1em plus 0.5em minus 0.4em\relax PMLR, 2021, pp. 1430--1440.

\bibitem{co2018self}
J.~Co-Reyes, Y.~Liu, A.~Gupta, B.~Eysenbach, P.~Abbeel, and S.~Levine, ``Self-consistent trajectory autoencoder: Hierarchical reinforcement learning with trajectory embeddings,'' in \emph{International conference on machine learning}.\hskip 1em plus 0.5em minus 0.4em\relax PMLR, 2018, pp. 1009--1018.

\bibitem{nachum2018near}
O.~Nachum, S.~Gu, H.~Lee, and S.~Levine, ``Near-optimal representation learning for hierarchical reinforcement learning,'' \emph{arXiv preprint arXiv:1810.01257}, 2018.

\bibitem{nasiriany2019planning}
S.~Nasiriany, V.~Pong, S.~Lin, and S.~Levine, ``Planning with goal-conditioned policies,'' \emph{Advances in neural information processing systems}, vol.~32, 2019.

\bibitem{li2022pre}
S.~Li, X.~Puig, C.~Paxton, Y.~Du, C.~Wang, L.~Fan, T.~Chen, D.-A. Huang, E.~Aky{\"u}rek, A.~Anandkumar \emph{et~al.}, ``Pre-trained language models for interactive decision-making,'' \emph{Advances in Neural Information Processing Systems}, vol.~35, pp. 31\,199--31\,212, 2022.

\bibitem{yao2023react}
S.~Yao, J.~Zhao, D.~Yu, N.~Du, I.~Shafran, K.~Narasimhan, and Y.~Cao, ``React: Synergizing reasoning and acting in language models,'' in \emph{International Conference on Learning Representations (ICLR)}, 2023.

\bibitem{hu2023look}
Y.~Hu, F.~Lin, T.~Zhang, L.~Yi, and Y.~Gao, ``Look before you leap: Unveiling the power of gpt-4v in robotic vision-language planning,'' \emph{arXiv preprint arXiv:2311.17842}, 2023.

\bibitem{belkhale2024rt}
S.~Belkhale, T.~Ding, T.~Xiao, P.~Sermanet, Q.~Vuong, J.~Tompson, Y.~Chebotar, D.~Dwibedi, and D.~Sadigh, ``Rt-h: Action hierarchies using language,'' \emph{arXiv preprint arXiv:2403.01823}, 2024.

\bibitem{nasiriany2024pivot}
S.~Nasiriany, F.~Xia, W.~Yu, T.~Xiao, J.~Liang, I.~Dasgupta, A.~Xie, D.~Driess, A.~Wahid, Z.~Xu \emph{et~al.}, ``Pivot: Iterative visual prompting elicits actionable knowledge for vlms,'' \emph{arXiv preprint arXiv:2402.07872}, 2024.

\bibitem{zhu2021hierarchical}
Y.~Zhu, J.~Tremblay, S.~Birchfield, and Y.~Zhu, ``Hierarchical planning for long-horizon manipulation with geometric and symbolic scene graphs,'' in \emph{2021 IEEE International Conference on Robotics and Automation (ICRA)}.\hskip 1em plus 0.5em minus 0.4em\relax Ieee, 2021, pp. 6541--6548.

\bibitem{jiao2022sequential}
Z.~Jiao, Y.~Niu, Z.~Zhang, S.-C. Zhu, Y.~Zhu, and H.~Liu, ``Sequential manipulation planning on scene graph,'' in \emph{2022 IEEE/RSJ International Conference on Intelligent Robots and Systems (IROS)}.\hskip 1em plus 0.5em minus 0.4em\relax IEEE, 2022, pp. 8203--8210.

\bibitem{sgbot_zhai2024sg}
G.~Zhai, X.~Cai, D.~Huang, Y.~Di, F.~Manhardt, F.~Tombari, N.~Navab, and B.~Busam, ``Sg-bot: Object rearrangement via coarse-to-fine robotic imagination on scene graphs,'' in \emph{2024 IEEE International Conference on Robotics and Automation (ICRA)}.\hskip 1em plus 0.5em minus 0.4em\relax IEEE, 2024, pp. 4303--4310.

\bibitem{grid_ni2024grid}
Z.~Ni, X.~Deng, C.~Tai, X.~Zhu, Q.~Xie, W.~Huang, X.~Wu, and L.~Zeng, ``Grid: Scene-graph-based instruction-driven robotic task planning,'' in \emph{2024 IEEE/RSJ International Conference on Intelligent Robots and Systems (IROS)}.\hskip 1em plus 0.5em minus 0.4em\relax IEEE, 2024, pp. 13\,765--13\,772.

\bibitem{LCBC_stepputtis2020language}
S.~Stepputtis, J.~Campbell, M.~Phielipp, S.~Lee, C.~Baral, and H.~Ben~Amor, ``Language-conditioned imitation learning for robot manipulation tasks,'' \emph{Advances in Neural Information Processing Systems}, vol.~33, pp. 13\,139--13\,150, 2020.

\bibitem{baku_haldar2024baku}
S.~Haldar, Z.~Peng, and L.~Pinto, ``Baku: An efficient transformer for multi-task policy learning,'' \emph{Advances in Neural Information Processing Systems}, vol.~37, pp. 141\,208--141\,239, 2024.

\bibitem{fang2024egocentric}
Z.~Fang, M.~Yang, W.~Zeng, B.~Li, J.~Yue, Z.~Ding, X.~Li, and Z.~Lu, ``Egocentric vision language planning,'' \emph{arXiv preprint arXiv:2408.05802}, 2024.

\bibitem{bousmalis2023robocat}
K.~Bousmalis, G.~Vezzani, D.~Rao, C.~Devin, A.~X. Lee, M.~Bauza, T.~Davchev, Y.~Zhou, A.~Gupta, A.~Raju \emph{et~al.}, ``Robocat: A self-improving foundation agent for robotic manipulation,'' \emph{arXiv preprint arXiv:2306.11706}, vol.~1, no.~8, 2023.

\bibitem{shah2023vint}
D.~Shah, A.~Sridhar, N.~Dashora, K.~Stachowicz, K.~Black, N.~Hirose, and S.~Levine, ``Vint: A foundation model for visual navigation,'' \emph{arXiv preprint arXiv:2306.14846}, 2023.

\bibitem{cot_wei2022chain}
J.~Wei, X.~Wang, D.~Schuurmans, M.~Bosma, F.~Xia, E.~Chi, Q.~V. Le, D.~Zhou \emph{et~al.}, ``Chain-of-thought prompting elicits reasoning in large language models,'' \emph{Advances in neural information processing systems}, vol.~35, pp. 24\,824--24\,837, 2022.

\bibitem{sam2_ravi2024sam}
N.~Ravi, V.~Gabeur, Y.-T. Hu, R.~Hu, C.~Ryali, T.~Ma, H.~Khedr, R.~R{\"a}dle, C.~Rolland, L.~Gustafson \emph{et~al.}, ``Sam 2: Segment anything in images and videos,'' \emph{arXiv preprint arXiv:2408.00714}, 2024.

\bibitem{mat_li2022mat}
W.~Li, Z.~Lin, K.~Zhou, L.~Qi, Y.~Wang, and J.~Jia, ``Mat: Mask-aware transformer for large hole image inpainting,'' in \emph{Proceedings of the IEEE/CVF conference on computer vision and pattern recognition}, 2022, pp. 10\,758--10\,768.

\bibitem{anydoor_chen2024anydoor}
X.~Chen, L.~Huang, Y.~Liu, Y.~Shen, D.~Zhao, and H.~Zhao, ``Anydoor: Zero-shot object-level image customization,'' in \emph{Proceedings of the IEEE/CVF conference on computer vision and pattern recognition}, 2024, pp. 6593--6602.

\bibitem{sgedit_zhang2024sgedit}
Z.~Zhang, D.~Chen, and J.~Liao, ``Sgedit: Bridging llm with text2image generative model for scene graph-based image editing,'' \emph{arXiv preprint arXiv:2410.11815}, 2024.

\bibitem{liu2024grounding}
S.~Liu, Z.~Zeng, T.~Ren, F.~Li, H.~Zhang, J.~Yang, Q.~Jiang, C.~Li, J.~Yang, H.~Su \emph{et~al.}, ``Grounding dino: Marrying dino with grounded pre-training for open-set object detection,'' in \emph{European conference on computer vision}.\hskip 1em plus 0.5em minus 0.4em\relax Springer, 2024, pp. 38--55.

\bibitem{yolo_redmon2016you}
J.~Redmon, S.~Divvala, R.~Girshick, and A.~Farhadi, ``You only look once: Unified, real-time object detection,'' in \emph{Proceedings of the IEEE conference on computer vision and pattern recognition}, 2016, pp. 779--788.

\bibitem{vaswani2017attention}
A.~Vaswani, N.~Shazeer, N.~Parmar, J.~Uszkoreit, L.~Jones, A.~N. Gomez, {\L}.~Kaiser, and I.~Polosukhin, ``Attention is all you need,'' \emph{Advances in neural information processing systems}, vol.~30, 2017.

\bibitem{hu2024unveiling}
X.~Hu, Z.~Xing, T.~Wang, C.-W. Fu, and P.-A. Heng, ``Unveiling deep shadows: A survey and benchmark on image and video shadow detection, removal, and generation in the deep learning era,'' \emph{arXiv preprint arXiv:2409.02108}, 2024.

\bibitem{act_zhao2023learning}
T.~Z. Zhao, V.~Kumar, S.~Levine, and C.~Finn, ``Learning fine-grained bimanual manipulation with low-cost hardware,'' \emph{arXiv preprint arXiv:2304.13705}, 2023.

\bibitem{uvd_zhang2024universal}
Z.~Zhang, Y.~Li, O.~Bastani, A.~Gupta, D.~Jayaraman, Y.~J. Ma, and L.~Weihs, ``Universal visual decomposer: Long-horizon manipulation made easy,'' in \emph{2024 IEEE International Conference on Robotics and Automation (ICRA)}.\hskip 1em plus 0.5em minus 0.4em\relax IEEE, 2024, pp. 6973--6980.

\bibitem{miles_papagiannis2024miles}
G.~Papagiannis and E.~Johns, ``Miles: Making imitation learning easy with self-supervision,'' \emph{arXiv preprint arXiv:2410.19693}, 2024.

\bibitem{dp_chi2023diffusion}
C.~Chi, Z.~Xu, S.~Feng, E.~Cousineau, Y.~Du, B.~Burchfiel, R.~Tedrake, and S.~Song, ``Diffusion policy: Visuomotor policy learning via action diffusion,'' \emph{The International Journal of Robotics Research}, p. 02783649241273668, 2023.

\bibitem{break-a-scene_avrahami2023break}
O.~Avrahami, K.~Aberman, O.~Fried, D.~Cohen-Or, and D.~Lischinski, ``Break-a-scene: Extracting multiple concepts from a single image,'' in \emph{SIGGRAPH Asia 2023 Conference Papers}, 2023, pp. 1--12.

\bibitem{lpips_zhang2018unreasonable}
R.~Zhang, P.~Isola, A.~A. Efros, E.~Shechtman, and O.~Wang, ``The unreasonable effectiveness of deep features as a perceptual metric,'' in \emph{Proceedings of the IEEE conference on computer vision and pattern recognition}, 2018, pp. 586--595.

\end{thebibliography}

\end{document}